\documentclass[conference]{IEEEtran}
\IEEEoverridecommandlockouts
\usepackage{cite}
\usepackage{amsmath,amssymb,amsfonts}
\usepackage{algorithmic}
\usepackage{graphicx}
\usepackage{float}
\usepackage{tabularray}
\usepackage{textcomp}
\usepackage{xcolor}
\def\BibTeX{{\rm B\kern-.05em{\sc i\kern-.025em b}\kern-.08em
    T\kern-.1667em\lower.7ex\hbox{E}\kern-.125emX}}
\makeatletter
\newcommand{\linebreakand}{%
  \end{@IEEEauthorhalign}
  \hfill\mbox{}\par
  \mbox{}\hfill\begin{@IEEEauthorhalign}
}
\makeatother
    
\begin{document}

\title{Place with Intention: An Empirical Attendance Predictive Study of Expo 2025 Osaka, Kansai, Japan}

\author{%
\IEEEauthorblockN{Xiaojie Yang}
\IEEEauthorblockA{\textit{The University of Tokyo}\\
Tokyo, Japan\\
xiaojieyang@g.ecc.u-tokyo.ac.jp}
\and
\IEEEauthorblockN{Dizhi Huang}
\IEEEauthorblockA{\textit{The University of Tokyo}\\
Tokyo, Japan\\
dizhi.huang@koshizuka-lab.org}
\and
\IEEEauthorblockN{Hangli Ge\textsuperscript{*}}
\IEEEauthorblockA{\textit{The University of Tokyo}\\
Tokyo, Japan\\
hanglige@g.ecc.u-tokyo.ac.jp}
\and
\linebreakand
\IEEEauthorblockN{Masahiro Sano}
\IEEEauthorblockA{\textit{Japan Association for the 2025 World Exposition
}\\
Osaka, Japan\\
SanoM@expo2025.or.jp}
\and
\IEEEauthorblockN{Takeaki Ohdake}
\IEEEauthorblockA{\textit{Japan Association for the 2025 World Exposition}\\
Osaka, Japan\\
OdakeT@expo2025.or.jp}
\linebreakand
\IEEEauthorblockN{Kazuma Hatano}
\IEEEauthorblockA{\textit{The University of Tokyo}\\
Tokyo, Japan\\
kazuma.hatano@koshizuka-lab.org}
\and
\IEEEauthorblockN{Noboru Koshizuka}
\IEEEauthorblockA{\textit{The University of Tokyo}\\
Tokyo, Japan\\
noboru@koshizuka-lab.org}
\thanks{\textsuperscript{*} Corresponding author.}%
}

\maketitle

\begin{abstract}
Accurate forecasting of daily attendance is vital for managing transportation, crowd flows, and services at large-scale international events such as Expo 2025 Osaka, Kansai, Japan. However, existing approaches often rely on multi-source external data (such as weather, traffic, and social media) to improve accuracy, which often leads to unreliable results because the available historical data are insufficient. To address these challenges, we propose a Transformer-based framework that leverages reservation dynamics, i.e., ticket bookings and subsequent updates within a time window, as a proxy for visitors’ attendance intentions, under the assumption that such intentions are eventually reflected in reservation patterns. This design avoids the complexity of multi-source integration while still capturing external influences like weather and promotions implicitly embedded in reservation dynamics. We construct a dataset combining entrance records and reservation dynamics and evaluate the model under both single-channel (total attendance) and two-channel (separated by East and West gates) settings. Results show that separately modeling East and West gates consistently improves accuracy, particularly for short- and medium-term horizons. Ablation studies further confirm the importance of the encoder–decoder structure, inverse-style embedding, and adaptive fusion module. Overall, our findings indicate that reservation dynamics offer a practical and informative foundation for attendance forecasting in large-scale international events.
\end{abstract}


\section{Introduction}

Accurate forecasting of daily attendance to public events is an essential research topic, with applications in urban planning, transportation management, sales forecasting, and event organization \cite{liu2019urban,rodrigues2016bayesian,lan2022research}. Public events can be examined in the context of large gatherings, such as concerts, sporting events, or theme parks, varying in duration, scale, and ticketing policies. In such large-scale gatherings, predicting crowd dynamics not only helps ensure public safety but also provides critical insights for resource allocation and infrastructure optimization. For instance, it directly influences decisions related to venue capacity, traffic control, and emergency preparedness \cite{ge2025traffic,ge2025simulation}, thereby mitigating the risks and inconveniences associated with overcrowding \cite{pang2024forecasting,zhang2017deep}.

Existing studies have explored attendance prediction in concerts \cite{rizi2019predicting}, sporting events \cite{arboretti2024predictive}, short-term festivals \cite{rizi2019predicting}, and long-term international expositions \cite{xie2000improving}. Many of these methods incorporate external variables such as weather, holidays, and transportation flows, which require large amounts of historical data to capture effective patterns. However, feature-rich approaches face several limitations: the collection and integration of diverse external data sources are costly, complex, and error-prone, and prediction accuracy does not always scale with the number of features, as irrelevant or noisy inputs can lead to unstable performance.

% The projected number of visitors is approximate 28.2 million in total, averaging more than 150,000 per day with even higher peaks on busy days. 

Against this background, Expo 2025 Osaka, Kansai, Japan presents unique challenges. Held in Osaka over a six-month period, it is one of the world’s largest international expositions, featuring a wide range of thematic pavilions, exhibitions, and cultural programs. The total number of visitors was approximately 29.0 million (25.6 million excluding Accreditation Pass holders), with a peak daily attendance exceeding 228,000 visitors. Accurate forecasting of daily attendance at Expo 2025 Osaka, Kansai, Japan is thus both a pressing operational need and a methodological challenge, for three reasons: (1) a stable predictive model is required from the outset of the six-month run for operations and safety management; (2) historical data are limited, and approaches that require extensive integration of heterogeneous external variables are impractical in this setting; and (3) frequent operational decisions and promotional campaigns significantly affect demand, yet such schedules are costly and difficult to model quantitatively. Moreover, visitors can modify their reservation dates multiple times, producing highly volatile reservation dynamics. Taken together, these factors make accurate attendance forecasting for Expo 2025 Osaka, Kansai, Japan both important and difficult.

To address these issues, we focus on the specific characteristics of the reservation system: making full use of the nature of visitor reservations. Our exploratory analysis shows that reservation dynamics inherently embed not only external shocks such as weather conditions and event announcements, but also visitors’ attendance intentions, as these intentions are expressed through booking and update behaviors. Building on this insight, we employ a Transformer-based time series forecasting model that learns directly from reservation dynamics, complemented by actual attendance records and calendar information. This design enables accurate and robust predictions without depending on complex multi-source data fusion.
% max value
% 周中，周末，花火大会，空演，恶劣天气（6.3）
% \begin{figure}
%     \centering
%     \includegraphics[width=1\linewidth]{610.png}
%     \caption{Enter Caption}
%     \label{fig1:intro_case}
% \end{figure}

The main contributions of this paper are summarized as follows:

\begin{itemize}
    \item We propose a Transformer-based framework for large-scale event attendance forecasting that leverages reservation dynamics as the primary predictive signal, thereby avoiding the complexity of multi-source data integration.
    \item We construct and analyze a dataset capturing multi-update reservation behaviors at Expo 2025 Osaka, Kansai, Japan, demonstrating its sensitivity to both external shocks and visitors’ attendance intentions.
    \item We conduct extensive experiments to validate the effectiveness of the proposed approach under different forecasting settings, and ablation studies further highlight the importance of key components such as the encoder–decoder structure and the adaptive fusion module.
\end{itemize}

\section{Related Works}
Time series forecasting has evolved considerably, moving from traditional statistical approaches to modern deep learning methods. Early studies mainly relied on classical techniques such as ARIMA \cite{box1976analysis}, exponential smoothing, and state space models, which provided interpretable frameworks but often struggled with nonlinear patterns and long-range dependencies. With the rise of machine learning, methods such as support vector regression and random forests were explored to capture more complex dynamics. More recently, deep learning has become the dominant paradigm, with recurrent neural networks (RNNs) \cite{elman1990finding}, long short-term memory (LSTM) networks \cite{hochreiter1997long}, and gated recurrent units (GRUs) \cite{cho2014learning} widely adopted for sequential data. However, these models often suffer from limitations in modeling long-term dependencies and scalability. To address this, Transformer-based architectures have emerged as a strong alternative, leveraging self-attention mechanisms to capture temporal correlations more effectively and enabling parallelized training. This shift has improved performance, particularly in tasks that require modeling long-range temporal dependencies.

Despite this progress, most existing research has concentrated on traffic flow forecasting in urban mobility and transportation systems \cite{kosugi2022traffic,hangli2022multi,matsunaga2023improving,ge2024k,ge2024frtp}, while studies specifically targeting large-scale event attendance remain limited. Nevertheless, accurate forecasting of daily or hourly visitor volumes is of critical importance for event organizers and city planners, as it enables effective crowd management \cite{anno2025early}, resource allocation \cite{awad2022event}, and safety measures \cite{sun2020predicting}. For instance, prior work on Expo 2010 Shanghai \cite{zhang2011daily} illustrates how predictive modeling can provide valuable operational insights for managing extraordinary demand surges during a world exposition. Similarly, studies leveraging social media signals \cite{de2017exploring} highlight the potential of user-generated content to infer attendance patterns, offering an alternative view into visitor behavior. Despite these promising efforts, research in this area is still relatively scarce compared to traffic forecasting, indicating a clear need for further exploration.

Another line of research has investigated the integration of diverse data sources to enhance forecasting accuracy. Prior studies have incorporated weather conditions, event calendars, and other external factors into predictive models, acknowledging that exogenous variables can strongly influence demand fluctuations. For example, ST-ResNet \cite{zhang2017deep} integrates external information such as weather, holidays, and events into deep spatio-temporal models for citywide crowd flow prediction. Similarly, Li et al. \cite{li2017forecasting} emphasize the role of event-related features in improving short-term passenger flow forecasts under unusual demand conditions. In addition, Liu et al. \cite{liu2019deeppf} demonstrate the effectiveness of combining multiple heterogeneous inputs to enhance predictive performance in public transit systems. However, these approaches often face two key challenges: first, the availability of high-quality and fine-grained external data is limited; second, preprocessing and aligning such heterogeneous sources introduces significant complexity.

In summary, existing research has either focused on transportation-related scenarios or relied heavily on complex integration of external variables, while applications to large-scale international events remain limited. At the same time, current approaches rarely model visitors’ attendance intentions explicitly \cite{chen2020dualsin,yang2025causalmob}. Yet intention-related information can be valuable in human mobility forecasting tasks, as it reflects not only external shocks such as weather or event announcements, but also the underlying motivation of visitors. Motivated by this, we consider reservation dynamics as a proxy for attendance intentions, providing a practical and informative basis for forecasting attendance in large-scale events.

\section{Dataset and Analysis}
In this study, we use two main sources of data provided by the Japan Association for the 2025 World Exposition: (i) historical attendance records aggregated at an hourly level, and (ii) ticket reservation data grouped into five predefined daily time slots. These datasets reflect both the observed visitor flows and the reservation patterns that precede them. Following our data handling policy, raw values and precise counts are not disclosed; instead, the dataset spans several consecutive months of operation and contains millions of records, sufficient to support robust model training and evaluation. 

The dataset consists of two core components: actual entrance records and reservation dynamics. Visitors are generally required to obtain tickets in advance, and reservations may be modified multiple times. These dynamic patterns not only reflect attendance intentions but also implicitly embed external influences such as weather forecasts and event announcements, making reservation dynamics a practical and informative proxy for forecasting attendance in large-scale events.

\subsection{Entrance Data}
The entrance dataset provides detailed logs of actual visitor admissions on an individual-ticket basis. Each record includes the date and time of entry (month, day, hour, and minute), the Reservation entry gate (East or West), the visitor’s country or regional origin, and the ticket attributes (type, description, and purchase channel). This dataset enables the aggregation of actual daily visitor counts and facilitates comparisons against pre-booked reservations.

\begin{figure}
    \centering
    \includegraphics[width=1\linewidth]{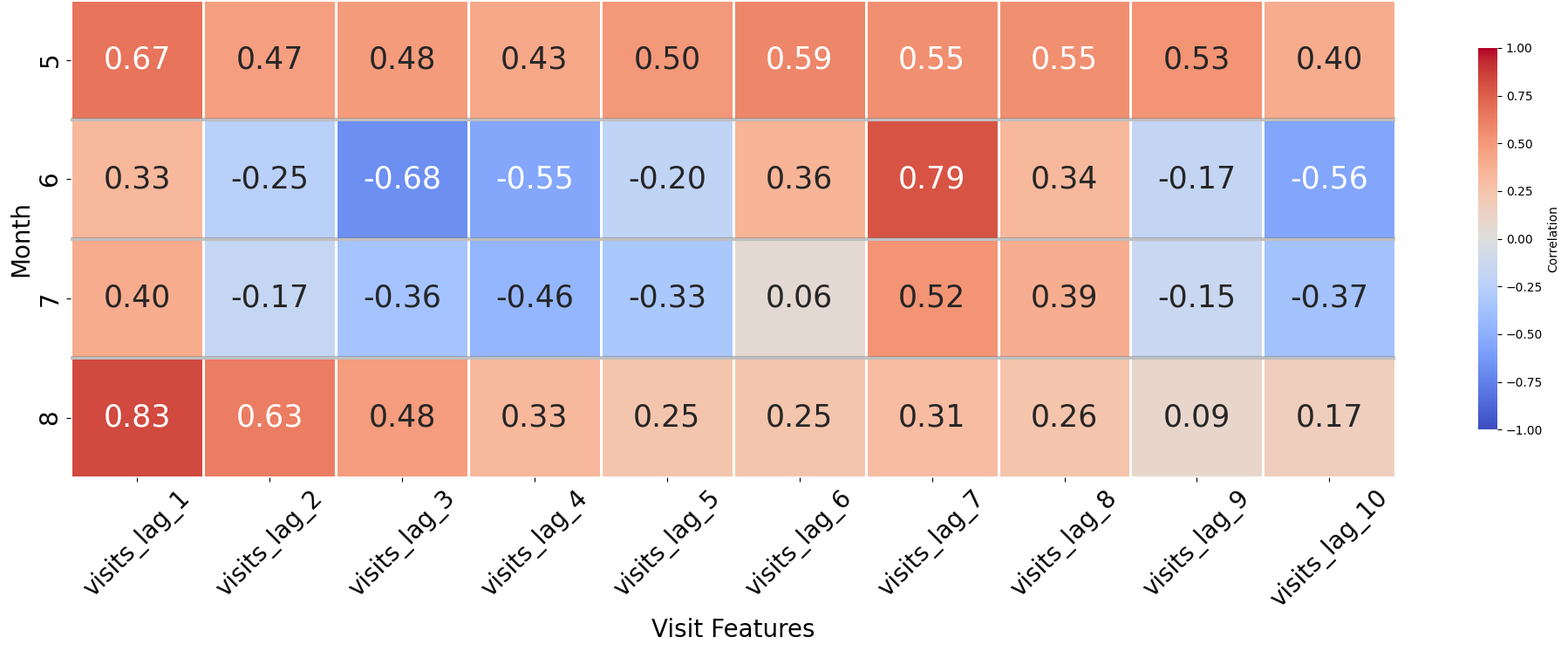}
    \caption{Heatmap of correlations between daily visits and lagged visits (1–10 days) across months (May–August). The figure shows that, besides the strong correlation with the previous day (lag 1), the 7-day lag generally exhibits the second-highest correlation, indicating a clear weekly periodicity in visit patterns}
    \label{fig:1}
\end{figure}

\subsection{Reservation Data}

The reservation dataset contains snapshots of ticket bookings. Each entry specifies the reservation date, gate category, and scheduled admission time, along with the available reservation capacity for the slot. Since reservations may be changed multiple times by the same visitor, the dataset provides a temporal series of reservation updates, which we term \textit{reservation dynamics}. These reservation updates help reveal how visitors adjust their plans over time.
\begin{figure}
    \centering
    \includegraphics[width=1\linewidth]{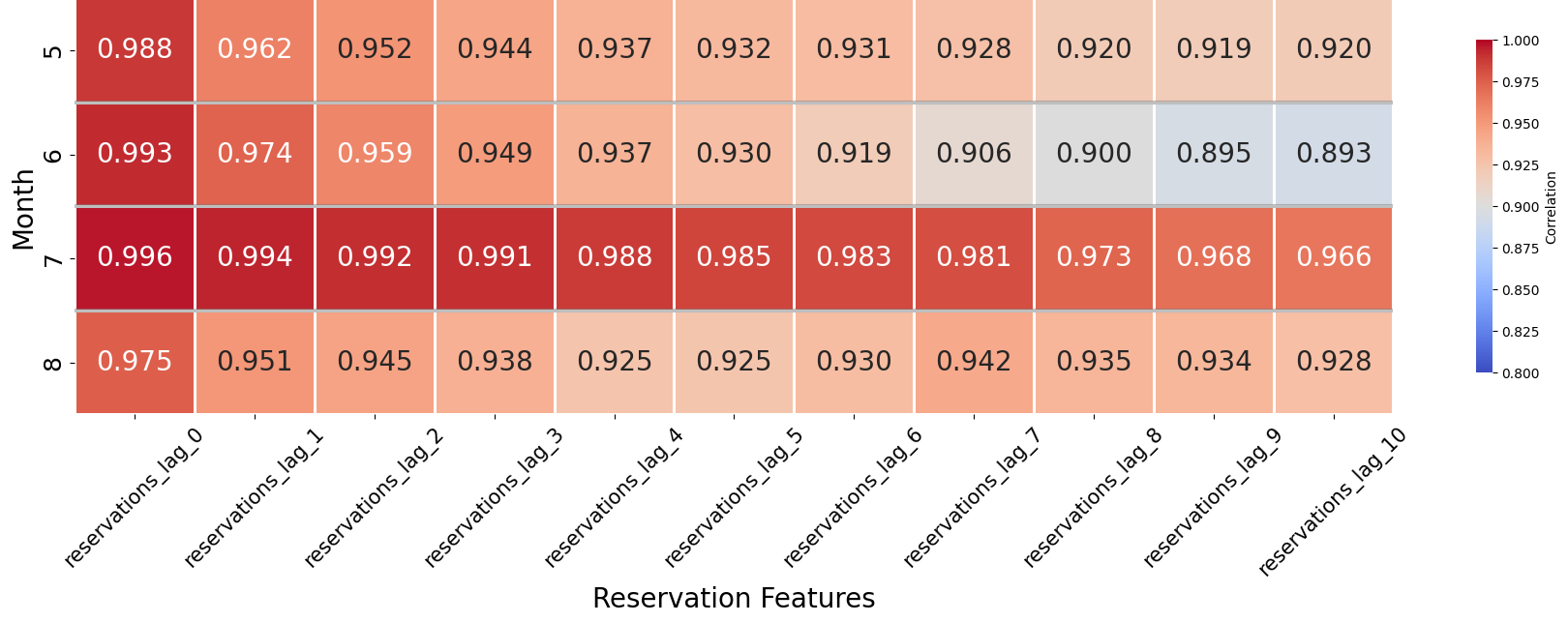}
    \caption{Heatmap of correlations between daily visit counts and reservation counts from up to 10 days earlier (lag 0–10), computed across months (May–August). The correlations are generally very high and increase as the reservation date approaches the actual visit day, highlighting the strong predictive relationship between reservations and real attendance}
    \label{fig:2}
\end{figure}

\begin{figure}
    \centering
    \includegraphics[width=1\linewidth]{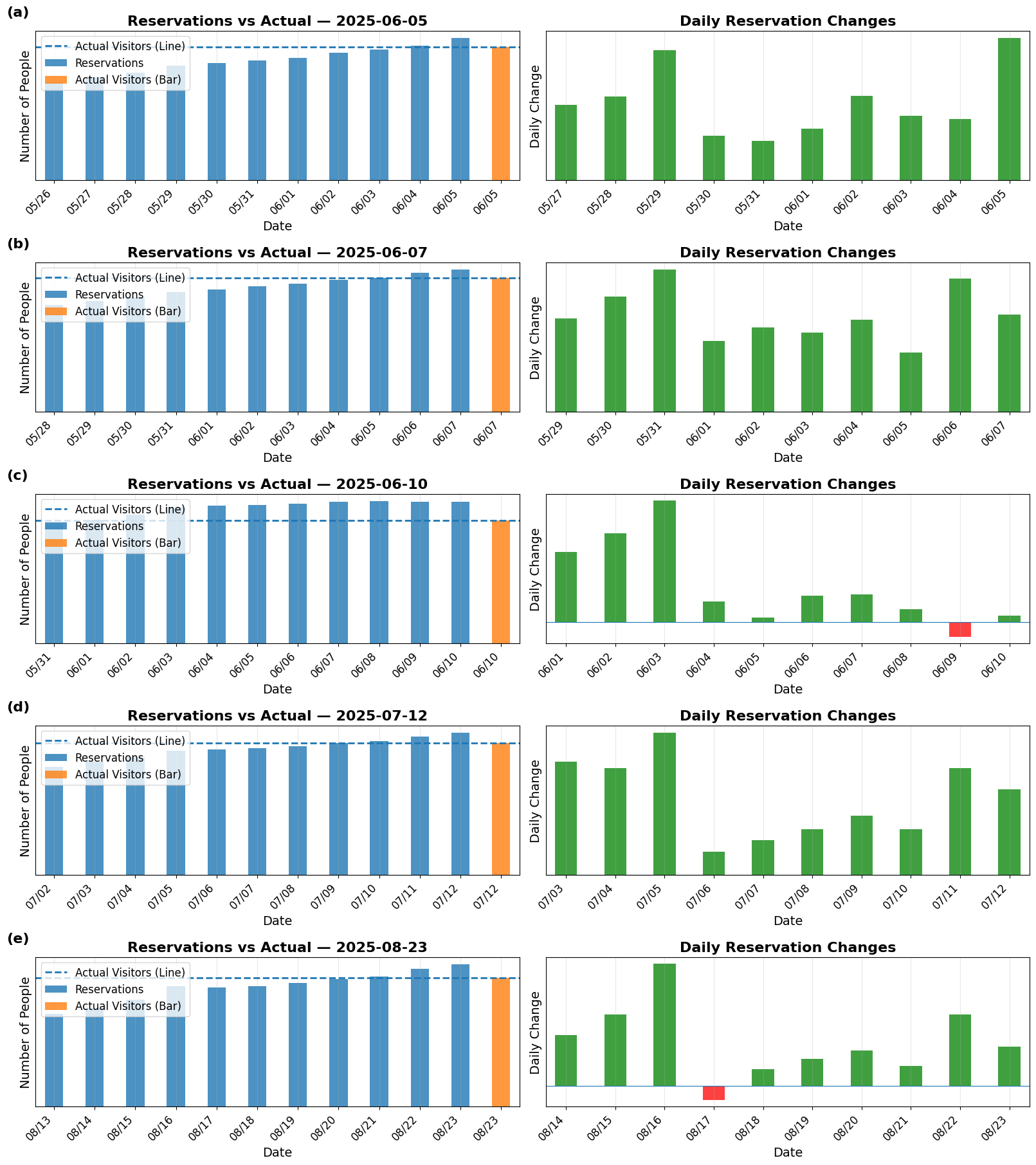}
    \caption{Case studies of reservation dynamics and daily changes for different dates. Subplots illustrate: (a) a regular weekday (June 5), (b) a weekend (June 7), (c) a rainy day (June 10), (d) the Blue Impulse air show day (July 12), and (e) a fireworks festival day (August 23). The left panels compare cumulative reservations with actual attendance, while the right panels depict daily reservation changes, highlighting how weather and special events strongly influence booking behavior.}
    \label{fig:case}
\end{figure}

\subsection{Analysis and Summary}

% \subsection{Summary}
To quantitatively assess the predictive value of historical and forward-looking features, we conducted a correlation analysis between daily ground truth attendance and two distinct categories of predictors: lagged visit counts (Visits\_lag) and lagged reservation updates (Reservations\_lag). Figure ~\ref{fig:case} presents the monthly correlation heatmaps for the period May–August.

The results reveal two complementary patterns. First, lagged visit counts exhibit moderate correlations with subsequent attendance, with the strongest associations observed at lag-1 and lag-7. This indicates that actual visitor flows are influenced both by the immediately preceding day and by the same weekday from the prior week. Such findings point to the coexistence of short-term temporal continuity and a recurring weekly cycle in visitor behavior. In contrast, correlations for more distant lags are substantially weaker or negative, underscoring the limited predictive utility of historical counts beyond these intervals.

Second, reservation dynamics demonstrate markedly stronger and more stable associations with real attendance. Across all months, the correlations consistently exceed 0.90, with the highest values observed for near-term lags (lag-0 and lag-1). This confirms that reservation updates function as a highly reliable proxy for actual attendance, as booking activity directly reflects real entries while implicitly integrating exogenous influences such as adverse weather or special event announcements through visitor modifications.

Overall, these results highlight two key empirical insights. On the one hand, entrance records capture short-term persistence and weekly periodicity in visitation patterns. On the other hand, reservation dynamics provide a superior and dynamic predictive signal that aligns closely with real attendance. These observations form the empirical foundation for the predictive modeling framework introduced in the subsequent section.

\section{Methodology}
\begin{figure*}[htbp]
\begin{center}
\includegraphics[scale=0.6]{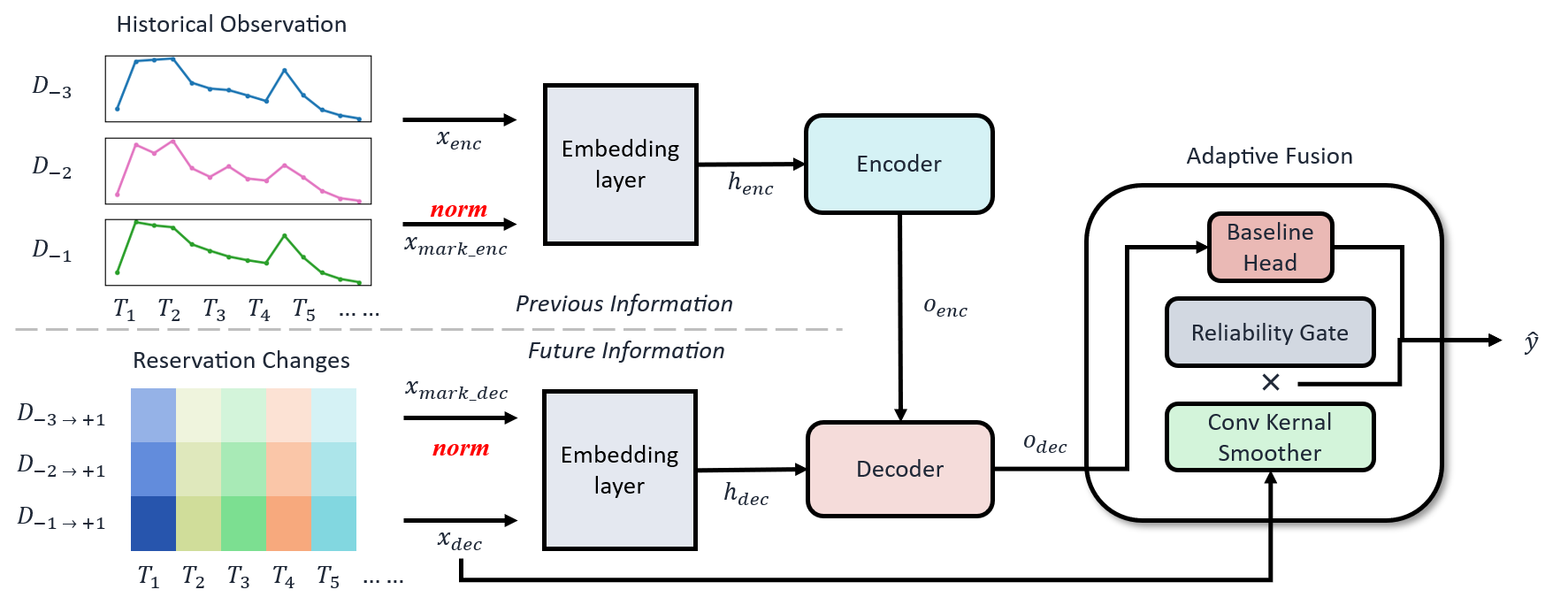}
\end{center}
\caption{
Model architecture overview. The encoder processes historical observations to extract contextual features $o_{\text{enc}}$, while the decoder integrates future-aware inputs and temporal markers to generate $o_{\text{dec}}$. The decoder output is then passed to the Adaptive Fusion module for final prediction $\hat{y}$.
}
\label{conceptual}
\end{figure*}
Building on the Transformer-based framework \cite{liu2023itransformer}, we design and evaluate two architectures for daily attendance forecasting: an \textit{encoder–decoder} model leveraging cross-attention over multiple pieces of information (entrance records and reservation dynamics) and a \textit{decoder-only} variant focusing solely on target-day reservation dynamics. To address the mismatch between reservation patterns and actual attendance, we introduce an adaptive fusion layer \cite{shao2022long}, which decomposes the prediction into three components: a nonnegative baseline, a learnable temporal kernel, and a reliability gate that modulates the contribution of reservation signals. Temporal features are embedded via linear projections. In the following, we will show more details about the data layout, embedding strategy, model architectures, and the adaptive fusion.

\subsection{Problem Setup}
We consider forecasting real attendance at \textit{Expo 2025 Osaka, Kansai, Japan} using three-dimensional temporal data. The information we have will be regarded as input:
\begin{align}
\mathbf{x}_{\text{enc}}       &\in \mathbb{R}^{(D_{\text{in}} \times T) \times C_{\text{real}}}, \quad
\mathbf{x}_{\text{enc\_mark}} \in \mathbb{R}^{(D_{\text{in}} \times T) \times d_{\text{temp}}}, \nonumber \\
\mathbf{x}_{\text{dec}}       &\in \mathbb{R}^{(D_{\text{out}} \times T) \times C_{\text{res}}}, \quad
\mathbf{x}_{\text{dec\_mark}} \in \mathbb{R}^{(D_{\text{out}} \times T) \times d_{\text{temp}}},
\end{align}

where $D_{\text{in}}$ is the number of historical days, $T$ is the number of hourly slots per day (e.g., 14), $C_{\text{real}}$ and $C_{\text{res}}$ denote the number of channels for real entrance records and reservation dynamics, respectively. The goal is to predict:
\begin{equation}
\hat{\mathbf{Y}} \in \mathbb{R}^{(D_{\text{out}} \times T) \times C_{out}},
\end{equation}

representing next $D_{\text{out}}$ days predicted entrance records across $T$ time slots. We focus on two situations where $C_{out}=1$ refers to the total number and $C_{out}=2$ means we output the number of East and West reserved entry gates.
\subsection{Temporal Embedding}

Each timestamp is annotated with four calendar fields: hour of day ($\in [0,13]$, from 8:00 am to 10:00 pm), weekday ($\in [0,6]$), month ($\in [0,12]$), and day of month ($\in [0,32]$). The dates and months in this work take leap years into account. Although the time span in our study is not large, we retain all temporal features in order to distinguish continuous patterns of change. Then, these integer-valued temporal features are normalized to $[0,1]$ as follows:
\begin{equation}
\tilde{\mathbf{x}}_{\text{mark}} =
\left[
\frac{\mathrm{hour}}{13},\;
\frac{\mathrm{weekday}}{6},\;
\frac{\mathrm{month}}{12},\;
\frac{\mathrm{day}}{32}
\right] \in [0,1]^4,
\end{equation}

where $\tilde{\mathbf{x}}_{\text{mark}}$ includes input information of encoder and decoder modules represented as $\mathbf{x}_{\text{enc\_mark}}$ and $\mathbf{x}_{\text{dec\_mark}}$.

Let $d$ denote the model latent tensors' dimension. Numeric values and temporal features are projected into a latent space via learned linear maps:
\begin{align}
\mathbf{z} &= \mathrm{concat}(\mathbf{x}, \mathbf{x}_{mark})\\
\mathbf{h} &= \mathrm{Dropout}\left( \mathrm{Linear}(\mathbf{z}^{T}) \right),
\end{align}
where $\mathbf{x}$ refers to input for encoder and decoder, and the input dimension of the Linear layer is $D \times T$, which is the total number of time steps for each prediction. We transpose the input so that each time step is treated as an individual unit of representation, enabling the model to better capture temporal dependencies across the entire sequence. This temporal embedding module is shared across encoder and decoder modules, resulting in:

\begin{equation}
\begin{aligned}
\mathbf{h}_{\text{enc}} &= \mathrm{TempEmbed}_{enc}(\mathbf{x}_{\text{enc}},\; \mathbf{x}_{\text{enc\_mark}}), \\
\mathbf{h}_{\text{dec}} &= \mathrm{TempEmbed}_{dec}(\mathbf{x}_{\text{dec}},\; \mathbf{x}_{\text{dec\_mark}}).
\end{aligned}
\end{equation}

\subsection{Model Architecture}
Given the representations $h_{\text{enc}} \in \mathbb{R}^{(C_{enc} + d_{temp}) \times d}$ and $h_{\text{dec}} \in \mathbb{R}^{(C_{enc} + d_{temp}) \times d}$ obtained from the shared temporal embedding module, we model temporal dependencies using a Transformer-based backbone \cite{vaswani2023attentionneed}.

When historical context is enabled, the encoder representation is processed via a stack of Transformer encoder layers to capture long-range and multi-scale patterns:
\begin{equation}
o_{\text{enc}} = \mathrm{TransformerEncoder}(h_{\text{enc}}) .
\end{equation}

The decoder representation attends both to itself and to the encoded memory. Specifically, it passes through masked self-attention layers to extract intra-period dependencies, followed by cross-attention layers to retrieve relevant information from the encoder output $o_{\text{enc}}$:
\begin{equation}
o_{\text{dec}} = \mathrm{TransformerDecoder}(h_{\text{dec}},\; o_{\text{enc}})
\end{equation}

In cases of decoder-only structure where historical context is not available, the encoder is removed and $h_{\text{dec}}$ is directly fed into a self-attentive decoder layer:
\begin{equation}
o_{\text{dec}} = \mathrm{TransformerDecoder}(h_{\text{dec}})
\end{equation}

In either case, $o_{\text{dec}}$ serves as the unified output representation for downstream forecasting tasks.

\subsection{Adaptive Fusion}

To transform the unified decoder representation $o_{\mathrm{dec}} \in \mathbb{R}^{L_{\mathrm{dec}} \times d}$ into final attendance predictions, we introduce an adaptive fusion mechanism that integrates a learned baseline with a temporally aligned reservation signal. For each time step $h=1,\dots,L_{\mathrm{dec}}$, the prediction is given by
\begin{equation}
\begin{aligned}
b_h &= \mathrm{softplus}\!\Big(\mathrm{MLP}_b(o_{\mathrm{dec},h})\Big),\\[4pt]
r_h &= \sigma\!\Big(\mathrm{MLP}_r(o_{\mathrm{dec},h})\Big),\\[4pt]
a^{(1)}_h &= \big(\mathbf{x}_{\mathrm{dec}}\,\mathbf{W}_{\mathrm{res}}\big)_h,\\[4pt]
\hat{y}_h &= b_h + r_h \sum_{u=-\lfloor K/2\rfloor}^{\lfloor K/2\rfloor} k_u\,a^{(1)}_{h-u}\,.
\end{aligned}
\end{equation}
where $o{\mathrm{dec},h}\in\mathbb{R}^d$ denotes the $h$-th decoder vector, $\mathrm{MLP}b$ and $\mathrm{MLP}r$ are multilayer perceptrons mapping $\mathbb{R}^d$ to $\mathbb{R}^{C{\text{out}}}$ that produce the nonnegative baseline $b_h$ and the gate value $r_h$, respectively. The term $\mathbf{x}{\mathrm{dec}}\in\mathbb{R}^{L_{\mathrm{dec}}\times C_{\mathrm{res}}}$ represents decoder inputs containing reservation dynamics, which are linearly aggregated by $\mathbf{W}{\mathrm{res}}\in\mathbb{R}^{C{\mathrm{res}}\times 1}$ into $a^{(1)}\in\mathbb{R}^{L_{\mathrm{dec}}}$. The kernel $k\in\mathbb{R}^{K}$ is normalized by softmax and applies a one-dimensional convolution over $a^{(1)}$ with same padding.

Our model separates the prediction into three components for interpretability: a stable baseline $b_h$, a temporally smoothed reservation signal parameterized by $k$, and a reliability gate $r_h$ that adaptively modulates the contribution of reservation information. Collecting all $\hat{y}h$ yields $\hat{\mathbf{y}} \in \mathbb{R}^{L{\mathrm{dec}} \times C_{\text{out}}}$ with $L_{\mathrm{dec}}=D_{\mathrm{out}}\times T$, which is then reshaped into $\hat{y} \in \mathbb{R}^{(D_{\mathrm{out}}\times T)\times C_{\text{out}}}$ and passed to the loss function.

\subsection{Objective Function}
The model is trained to minimize hourly prediction error over each target day using the mean absolute error (MAE). Let $\hat{y}, y \in \mathbb{R}^{L \times C_{out}}$ denote the predicted and ground-truth attendance curves for a single example, where $L$ is the number of hourly slots. The loss is defined as:
\begin{equation}
\label{eq:loss}
\mathcal{L} =
\frac{1}{L}\sum_{h=1}^{L} \lvert \hat{y}_h - y_h \rvert,
\end{equation}
where $\hat{y}$ is the predicted number and $y$ is the ground truth.

This objective directly supervises the predicted attendance values at each time slot and serves as the sole training loss.

\section{Experiments}
\subsection{Experiment Settings}
% The dataset is divided into training, validation, and test sets with a ratio of 0.7/0.1/0.2. The best hyperparameters are determined based on the validation set using k-fold cross-validation. Each test utilizes a uniform model structure, maintaining the same number of parameters. In conventional models, two travel patterns are merged together for training, whereas for causality-aware prediction model, nonanchor targeted travels are implemented with counterfactuals as causal inference for training as we illustrated in Fig. \ref{framework}. Embedding dimensions are set to 128 for the user, location, and time information and 256 for the final hidden state if the original paper do not provide detailed information. The experiments are conducted on a dl-box GPU server with four NVIDIA RTX A6000 graphic cards, using Python 3.9, PyTorch 1.9, and Cuda Toolkit 11.7.
We use the reservation-update logs (\emph{reservation dynamics}) and the real-world entrance records provided by the Japan Association for the 2025 World Exposition, from 23rd April to 17th September, aggregating both into hour-level slots (14 slots per day from 08:00 to 22:00). Time-ordered samples are split chronologically into training and test spans with an 80\%/20\% ratio. We choose hyperparameters based on grid search and early stopping on the training set. To keep the model compact and stable under limited historical data, we use the following configuration: 16-dimensional embeddings and hidden states, a single Transformer layer for the encoder and the decoder, and a dropout rate of 0.1. We train with the Adam optimizer (learning rate $1\times10^{-3}$) using mini-batches of size 4. Unless otherwise noted, the random seed is fixed at 3407 for reproducibility. Experiments are conducted on Ubuntu 18.04.6 with NVIDIA RTX A6000 (48\,GB) GPUs, CUDA 12.1, Python 3.11.13,  and PyTorch 2.4.0.

\subsection{Overall Performance}
\begin{table*}[!htbp]
\centering
\renewcommand{\arraystretch}{1.3} % 行距调节
\caption{Prediction performance under the \textbf{single-channel} setting, where the model forecasts the \textit{total daily attendance} (aggregated over East and West gates). Each cell reports MAE / MAPE.}
\begin{tabular}{c|c|c|c|c|c}
\hline
\textbf{Input window $\backslash$ Horizon} & \textbf{1 Day} & \textbf{3 Days} & \textbf{5 Days} & \textbf{7 Days} & \textbf{14 Days} \\
\hline
1 Day  & 1297.37 / 2898.36 & 1246.31 / 3194.99 & 2031.08 / 165757.14 & 1449.90 / 358605.16 & 2002.21 / 524438.06 \\
3 Days & 1648.31 / 3011.06 & 1215.83 / 2937.72 & 1271.20 / 174412.03 & 1338.39 / 253493.75 & 1834.44 / 556289.13 \\
5 Days & 1218.67 / 2480.13 & 1292.49 / 3166.85 & 1355.42 / 2767.44 & 1350.32 / 261855.47 & 1686.71 / 488265.38 \\
7 Days & 1114.37 / 4333.98 & 1229.06 / 2764.93 & 1166.66 / 2738.09 & 1420.83 / 136018.89 & 1767.81 / 480190.59 \\
14 Days & 1474.29 / 1576.90 & 1293.33 / 2826.56 & 1285.90 / 2628.25 & 1346.23 / 2255.03 & 1523.61 / 440984.03 \\
\hline
\end{tabular}
\label{tab:single}
\end{table*}

\begin{table*}[!htbp]
\centering
\renewcommand{\arraystretch}{1.3} % 行距调节
\caption{Prediction performance under the \textbf{two-channel} setting, where the model forecasts \textit{East and West gate attendances separately}, which are then aggregated. Each cell reports MAE / MAPE.}
\begin{tabular}{c|c|c|c|c|c}
\hline
\textbf{Input window $\backslash$ Horizon} & \textbf{1 Day} & \textbf{3 Days} & \textbf{5 Days} & \textbf{7 Days} & \textbf{14 Days} \\
\hline
1 Day & 733.46 / 1622.46 & 852.02 / 2500.81 & 1001.48 / 48240.03 & 875.39 / 100293.66 & 1208.29 / 140094.03 \\
3 Days & 714.62 / 1443.48 & 829.28 / 1151.04 & 843.96 / 48199.61 & 830.62 / 71973.78  & 1038.93 / 145116.22 \\
5 Days & 914.34 / 2029.01 & 774.46 / 1092.18 & 886.68 / 2419.30  & 837.05 / 70301.27 & 971.87 / 134340.58 \\
7 Days & 816.88 / 1809.91 & 820.30 / 2201.54 & 745.86 / 1944.82  & 858.65 / 38448.59 & 964.98 / 133560.67 \\
14 Days & 842.82 / 1311.30 & 853.64 / 2217.00 & 855.97 / 1880.95  & 912.79 / 2296.74  & 1082.87 / 120917.18 \\
\hline
\end{tabular}
\label{tab:multi}
\end{table*}

\begin{table}[!htbp]
\centering
\renewcommand{\arraystretch}{1.3} % 行距调节
\caption{Prediction performance comparison with baselines under the \textbf{two-channel} setting, where experiements focuses the 7 days input window and 5 days horizon. Each cell reports MAE / MAPE.}
\begin{tabular}{c|c}
\hline
Models & MAE / MAPE \\
\hline
ARIMA & 1170.59 / 3695.75 \\
LSTM & 1118.51 / 3430.64  \\
GRU  & 954.21 / 3143.72 \\
TCN  & 779.10 / 3273.98  \\
Ours & \textbf{745.86} / \textbf{1944.82}  \\
\hline
\end{tabular}
\label{tab:baseline}
\end{table}

We evaluate forecasting accuracy using two standard error metrics: mean absolute error (MAE) and mean absolute percentage error (MAPE), defined respectively as
\begin{equation}
\mathrm{MAE} = \frac{1}{N} \sum_{i=1}^{N} \left| y_i - \hat{y}_i \right|,
\end{equation}
\begin{equation}
\mathrm{MAPE} = \frac{1}{N} \sum_{i=1}^{N} \left| \frac{y_i - \hat{y}_i}{y_i} \right|,
\end{equation}
where $\hat{y}$ is the predicted number and $y$ is the ground truth.

The prediction results under the single-channel setting, where the model directly predicts the aggregated number of hourly visitors, are reported in Table~\ref{tab:single}. In contrast, Table~\ref{tab:multi} presents the outcomes of the two-channel setting, in which human flows of the East and West reserved entry gate are modeled separately. The comparison highlights that the two-channel approach consistently achieves lower MAE values, particularly for shorter horizons. For example, when the input window is 3 and the prediction horizon is 1, the MAE decreases from 1648.31 in Table~\ref{tab:single} to 714.62 in Table~\ref{tab:multi}, with a corresponding reduction in MAPE. Similarly, at a horizon of 3 days, the two-channel setting maintains more stable errors, with MAE around 829.28 compared to 1215.83 in the single-channel case.

To further compare the proposed model, we compare it with several widely adopted time-series forecasting baselines, including statistical and neural architectures, as follows:

\begin{itemize}

    \item \textbf{ARIMA\cite{williams2003modeling}}: Utilizes autoregressive and moving‐average components to capture short- and medium-range temporal dependencies in a single series, offering simplicity but limited multivariate flexibility.

    \item \textbf{LSTM\cite{hochreiter1997long}}: Introduces memory cells and gating mechanisms to mitigate vanishing and exploding gradients, thereby effectively capturing long-term temporal dependencies and outperforming vanilla RNNs in sequential modeling and prediction stability.
    
    \item \textbf{GRU\cite{chung2014empirical}}: Employs gated update and reset mechanisms to model sequential dependencies, matching LSTM performance with fewer parameters and faster training.
    
    \item \textbf{TCN\cite{bai2018empirical}}: Employs stacked one-dimensional causal and dilated convolutions with residual connections to capture long-range temporal dependencies efficiently, enabling parallel computation, stable gradients, and faster convergence compared to recurrent architectures.

\end{itemize}

In Table~\ref{tab:baseline} we report baseline and proposed model performance in the two channel setting with a 7 day input window and a 5 day horizon. Using MAE and MAPE, our model attains the best accuracy with 745.86 MAE and 1944.82 MAPE. Compared with the strongest baseline TCN, errors drop by 5.9\% and 44.9\%, respectively, showing clear gains from learning cross gate dependencies and stabilizing multi channel forecasts. In contrast, RNN based and convolutional baselines handle smoother weekly cycles but adapt poorly to abrupt reservation updates and inter gate interactions in this dataset. The Transformer with self attention can attend to irregular yet correlated shifts in reservation signals and entrance records, which explains the observed advantage.

The results show that modeling the two gates separately leads to more accurate predictions. Furthermore, moderate input windows (e.g., IW=5 or IW=7) tend to obtain the most favorable trade-off between information richness and generalization across horizons. For longer horizons such as ID=14, both settings exhibit larger errors, and MAPE values can become inflated due to very small denominators, despite MAE remaining within a reasonable range. Overall, the two-channel prediction provides more reliable and robust forecasts of daily attendance at Expo 2025 Osaka, Kansai, Japan.

\subsection{Ablation Study}
Table~\ref{tab:ablation} reports the results of the ablation experiments. Here, \textit{w/o Inv} denotes the removal of the inverse-style temporal embedding where variables are treated as tokens, \textit{DecOnly} corresponds to a decoder-only framework without the encoder or cross-attention that incorporates historical entrance records, \textit{w/o AF} removes the adaptive fusion module that integrates baseline prediction with reliability-gated kernel transformation, and \textit{DecOnly w/o AF} combines both simplifications. The full model (\textit{Full }) incorporates all components, including the encoder--decoder structure, inverse-style embedding, and adaptive fusion.
\begin{table}[!htbp]
\centering
\renewcommand{\arraystretch}{1.2}
\caption{Ablation study results under the 7-day input window and 5-day horizon setting. Each cell reports MAE / MAPE.}
\begin{tabular}{l|c}
\hline
\textbf{Variant} & \textbf{MAE / MAPE} \\
\hline
w/o Inv        & 1276.11 / 2504.85 \\
DecOnly w/o AF & 1898.91 / 2231.11 \\
DecOnly        & 1242.90 / 2597.35 \\
w/o AF         & 2608.83 / 1906.53 \\
Full    & \textbf{1166.66 / 2738.09} \\
\hline
\end{tabular}
\label{tab:ablation}
\end{table}

The results demonstrate that the adaptive fusion mechanism plays the most critical role, as its removal leads to a dramatic degradation in performance, with MAE rising to 2608.83. The encoder–decoder framework also provides clear advantages compared to the decoder-only setting, indicating that cross-attention with historical sequences is beneficial. Furthermore, the inverse-style embedding contributes consistent improvements: without it, performance declines to 1276.11 in MAE compared to the full model’s 1166.66. Overall, the full model achieves the best results, confirming the complementary contributions of the encoder–decoder architecture, inverse-style embedding, and adaptive fusion in enhancing predictive accuracy.

\section{Conclusion}
This paper addresses daily attendance forecasting for Expo 2025 Osaka, Kansai, Japan, using a lightweight Transformer-based framework that employs reservation dynamics as the main signal. As stated in the Introduction, this design avoids heavy multi-source fusion while capturing visitors’ intentions embedded in booking updates. Experiments under both single- and two-channel settings show that modeling the East and West reserved entry gates separately yields more accurate and stable forecasts, especially at short- and medium-term horizons. Ablation studies further highlight the complementary contributions of the adaptive fusion module, the encoder–decoder structure with cross-attention, and the inverse-style temporal embedding, each of which improves predictive performance in different ways. While the framework is practical and compact, it still relies on the stability of the reservation system and may face difficulties on days with sudden policy changes or rare events; its generalization beyond Expo 2025 Osaka, Kansai, Japan also remains open. Beyond daily operations, the forecasts can also assist in planning visitor flows and improving the on-site experience. All analyses use aggregated, anonymous reservation data under data minimization and access control, and intention inference is limited to capacity planning and safety, with commitments to transparency and periodic bias checks. Overall, the findings confirm that reservation dynamics can obtain reliable and robust forecasts, offering useful support for daily operations, resource allocation, safety management, and visitor wellbeing in large-scale international events.

\section*{Acknowledgment}

The authors would like to express sincere gratitude to all those who contributed to the completion of this study. In particular, we are deeply indebted to the members of the \textit{Data Utilisation Initiatives in Expo 2025 Osaka, Kansai, Japan} for their invaluable advice and generous support: \textbf{Atsuo KISHIMOTO} (Director \& Professor, The University of Osaka Research Centre on Ethical, Legal and Social Issues); \textbf{Shinji SHIMOJO} (Professor, Faculty of Software and Information Sciences, Aomori University and Professor Emeritus, The University of Osaka); \textbf{Daisuke TAKAYANAGI} (Director General of IT Security Center, Information-technology Promotion Agency, Japan (IPA)); \textbf{Hiroaki MIYATA} (Professor, Keio University School of Medicine); \textbf{Tatsuhiko YAMAMOTO} (Professor, Keio University Law School). Furthermore, the provision of meteorological data by \textbf{Japan Meteorological Corporation} made the execution of this research possible. We extend our deepest appreciation for the kindness and support of all those involved. The views and opinions expressed in this paper are those of the authors only and do not necessarily reflect those of their affiliated institutions, the funding agencies or acknowledged contributors.

\bibliographystyle{IEEEtran}
\bibliography{IEEEexample}

\end{document}